\documentclass[lettersize,journal]{IEEEtran}
\usepackage{amsmath,amsfonts}
\usepackage{algorithmic}
\usepackage{algorithm}
\usepackage{array}
\usepackage[caption=false,font=normalsize,labelfont=sf,textfont=sf]{subfig}
\usepackage{textcomp}
\usepackage{stfloats}
\usepackage{url}
\usepackage{verbatim}
\usepackage{graphicx}
\usepackage{cite}
\usepackage{tabularx}
\usepackage{multirow} 
\usepackage{makecell} 

\begin{document}

\title{Scene-Adaptive Motion Planning with Explicit Mixture of Experts and Interaction-Oriented Optimization}

\author{Hongbiao Zhu\IEEEauthorrefmark{1}, Liulong Ma\IEEEauthorrefmark{1}\IEEEauthorrefmark{2}, Xian Wu, Xin Deng, Xiaoyao Liang
\thanks{*These authors contributed equally to this work.}
\thanks{\IEEEauthorrefmark{2} indicates corresponding author. LiulongMa@outlook.com}
\thanks{This paper was produced by the Automotive New Technology Research Institute, BYD Company Limited. They are in Shenzhen, China.}}



\maketitle

\begin{abstract}
Despite over a decade of development, autonomous driving trajectory planning in complex urban environments continues to encounter significant challenges. These challenges include the difficulty in accommodating the multi-modal nature of trajectories, the limitations of the single expert model in managing diverse scenarios, and insufficient consideration of environmental interactions. To address these issues, this paper introduces the EMoE-Planner, which incorporates three innovative approaches. Firstly, the Explicit MoE (Mixture of Experts) dynamically selects specialized experts based on scenario-specific information through a shared scene router. Secondly, the planner utilizes scene-specific queries to provide multi-modal priors, directing the model’s focus towards relevant target areas. Lastly, it enhances the prediction model and loss calculation by considering the interactions between the ego vehicle and other agents, thereby significantly boosting planning performance. Comparative experiments were conducted on the Nuplan dataset against the state-of-the-art methods. The simulation results demonstrate that our model consistently outperforms SOTA models across nearly all test scenarios. Our model is the first pure learning model to achieve performance surpassing rule-based algorithms in almost all Nuplan closed-loop simulations. 
\end{abstract}


\section{Introduction}
\IEEEPARstart{A}{utonomous} driving trajectory planning has evolved over decades, with rule-based methods~\cite{rule1,rule2,rule3} providing fundamental safety assurances via predefined logic and heuristics. However, in complex urban settings, three significant limitations become apparent: (1) The manual construction of rules struggles to accommodate dynamic interactions and abrupt changes in road topology, resulting in unaddressed long-tail scenarios; (2) Rigid trajectory generation fails to mimic the adaptive behaviors of human drivers, such as dynamically adjusting following distances; (3) An exponential increase in maintenance costs arises from the “combinatorial explosion” of accumulating rules.\par

Conversely, data-driven approaches, including imitation learning~\cite{pluto,plantf,gameformer}, address edge cases like extreme weather and complex traffic, capturing human-like driving behaviors from expert data. Reinforcement learning~\cite{reinforce1,reinforce3} enables dynamic optimization through advanced reward mechanisms. These systems offer lower costs and faster iterations compared to rule-based alternatives.

Despite their benefits, two primary challenges persist in complex urban environments:
1) Urban driving demands highly multi-modal planning to address dynamic constraints and interactive intentions. Existing methods \cite{uniad,drivelm,pluto} generate multiple candidate trajectories but require post-processing modules to select the best modality. The complexity of scenarios like roundabouts and intersections further complicates this issue, with single-expert networks struggling to perform consistently across varied situations.
2) Traditional imitation learning employs static loss functions that neglect agent-ego interactions, leading to higher collision rates in dense traffic compared to rule-based methods. 

To address these challenges, this paper presents an explicit MoE model that incorporates ego-agent interactions, scene-specific queries for multi-modal planning in urban scenarios. The contributions of this paper include the following three points.
\begin{itemize}
\item Propose an Explicit Mixture of Experts (EMoE) that replaces the black-box routing in traditional MoE with a semantically meaningful scene classifier. This explicit routing mechanism assigns well-defined tasks to each expert, thereby reducing their individual learning complexity. Moreover, it ensures inherent load balance by allowing direct control over the training data distribution across different scenario types.
\item Introduce a scene-specific query that dynamically adapts to the scenario type identified by EMoE architecture. The scene-specific query confines the output space to only relevant maneuvers (e.g., left turns for a left-turn scenario), thereby raising the quality floor by avoiding the generation of unrealistic or poor-quality trajectories.
\item Introduce an Interaction-Oriented Loss to steer the model's attention toward what is more important within an interaction. Furthermore, the prediction of surrounding agents trajectories considers their interaction with the ego vehicle, which not only improves the prediction accuracy but also indirectly optimizes the scene encoding. \par
\end{itemize}

\section{Related Work}
\label{sec:related-work}
\paragraph*{Multi-modal modeling} Multi-modality is a significant and persistent challenge for autonomous driving trajectory planning. Early approaches~\cite{anchor-free1,transfuser} overlooked the inherent multi-modal nature of driving scenarios, particularly anchor-free methods that relied solely on environmental and map information to predict future trajectories. Even recent end-to-end models like UniAD~\cite{uniad}, specifically its motion former component, leverage map and object information to generate trajectories via anchor-free motion queries. However, these approaches risk modal collapse due to their limited capacity to represent diverse potential behaviors.
Current state-of-the-art models are increasingly adopting anchor-based strategies. PLUTO~\cite{pluto} employ a semi-anchor approach, leveraging reference lines as lateral anchors and anchor-free queries longitudinally to approximate multi-modality. Despite their impressive performance, these methods are limited by a dependence on detailed maps. Since perception models cannot yet reliably output reference lines directly, rule-based layers are still required, which complicates deployment and prevents full end-to-end learning. Alternatively, VAD2~\cite{vad2} and Hydra MDP~\cite{hydra} utilize pre-generated trajectory libraries as anchors to cover a wide range of possible trajectory modalities. These methods select the optimal trajectory by scoring candidates. Further refinements, such as applying Gaussian Mixture Models (GMMs) to trajectory generation~\cite{traj4,traj5} or sampling trajectories from predicted heatmaps~\cite{traj7,traj8}, aim to improve the distribution rationality and selection probability of generated trajectories. Diffusion models are gaining traction in trajectory generation~\cite{diffuser1,diffusiondrive,diffusionplanner} due to their inherent multi-modal capabilities. By dividing the action space with predefined anchor points and employing a truncated diffusion policy, DiffusionDrive~\cite{diffusiondrive} generates multimodal trajectories ten times faster than traditional diffusion strategies. However, the generated trajectories lack adequate constraints, making it challenging to balance high quality with diversity, and in some cases, it may even produce trajectories of very poor quality. To address this, DiffusionPlanner~\cite{diffusionplanner} augments a pre-trained diffusion model with classifier guidance, which applies constraints to the trajectories. This ensures both the quality and diversity of the generated paths while effectively avoiding low-quality results.\par

Consequently, researchers have explored using trajectory endpoints or destination points as anchors~\cite{tnt,mtr,goalflow}. MTR~\cite{mtr} directly generates trajectory endpoints by clustering ground truth data, leveraging endpoints as intent points with learnable position embeddings serving as static intent queries. GoalFlow~\cite{goalflow}, builds upon these approaches by directly generating trajectory endpoints as anchors.
While endpoint anchors enhance trajectory interpretability and preserve multi-modality, current methods use a single anchor set for all scenes. Dense anchor sets increase computation, while sparse anchor sets limit modality. This paper introduces a model with scene-specific endpoint anchors, balancing anchor density and modality representation across diverse driving environments. \par

Furthermore, the complexity of urban scenarios presents unique challenges, as differences between scenes persist despite some common planning rules. STR2~\cite{str2} address this by replacing the GPT2 backbone in STR~\cite{str} with an MoE architecture, substituting the feedforward network in the Transformer~\cite{transformer} with an MoE layer.  This adaptation leverages MoE’s classification capabilities, significantly boosting performance in closed-loop simulations.
MoE~\cite{moe1,moe2}, widely used in large language models (LLM) like deepseek-v2~\cite{deepseek}, Qwen1.5~\cite{qwen} and Switch Transformer~\cite{switchtransformer}, is apt for planning tasks that require scenario-specific strategies. However, traditional MoE’s unsupervised expert selection often leads to imbalanced loads and reduced interpretability. While techniques like auxiliary load balancing losses, as proposed in Switch Transformer~\cite{switchtransformer} and DeepSeek-V2~\cite{deepseek}, can mitigate this issue, they also distort the overall loss landscape and potentially compromise model generalization. Therefore, this paper introduces an improved MoE architecture tailored for trajectory planning, which is not only highly interpretable but also provides valuable insights into data collection strategies based on the performance of individual experts. \par

\paragraph*{Interactive game modeling} Modeling multi-agent interactions remains a critical challenge in autonomous driving trajectory planning. The mutual influence between the planned trajectory and predicted trajectories presents fundamental complexity. Prediction-first paradigms~\cite{pred-planning1,uniad,vad} sequentially predict agent trajectories then plan ego motion, but neglect the ego’s impact on others, often yielding over-conservative behaviors. Planning-first paradigms~\cite{planning-pred1,planning-pred2} conversely plan ego trajectories before predicting agent responses, risking over-aggressive maneuvers through unrealistic collision-avoidance assumptions. Joint optimization frameworks~\cite{union1,union2} simultaneously optimize both trajectories with shared objectives, yet suffer from prediction-execution mismatches when agent cooperation assumptions violate real-world responses. Iterative refinement~\cite{iter1,iter2}, while theoretically optimal for capturing interaction dynamics, incurs prohibitive computational costs for real-time applications. GameFormer~\cite{gameformer} employs a layered architecture in which each layer refines its result based on predictions from the previous layer, thereby balancing computational cost and performance by adjusting the number of layers. To fully consider the influence of interaction, this paper adapts planning-first paradigm and introduces an interactive regression loss that penalizes trajectory deviations caused by mutual ego-agent influences during optimization. \par

\section{Methodology}
EMoE-Planner is a model proposed to address the trajectory planning problem for autonomos driving in urban scenarios. This section provides a detailed introduction to the model, including the specific implementation and the underlying motivations for its design.

\subsection{Overview}
\begin{figure*}[t]
  \centering 
  \includegraphics[width=0.85\textwidth]{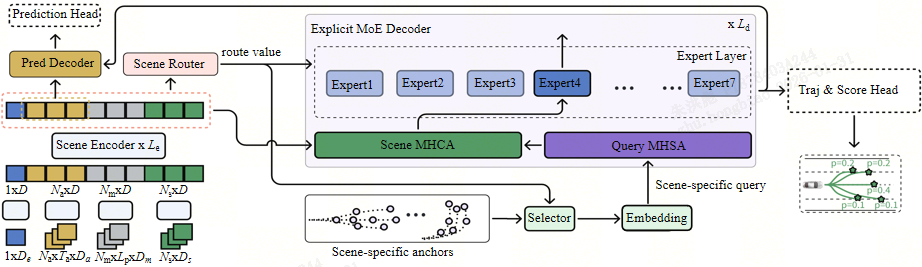} 
  \caption{Framework of EMoE-Planner} 
  \label{fig:framework}
  \vspace{-0.5cm}  
\end{figure*}

\paragraph*{Problem Description}
The trajectory planning task in autonomous driving aims to generate safe, comfortable and traffic law-compliant vehicle trajectories for a future time horizon $T_f$, given the ego vehicle’s state and surrounding environmental information, along with the navigation goal. Let the ego vehicle’s state be denoted as $F_e$. The surrounding environmental information includes dynamic objects ($F_a$) such as vehicles and pedestrians, static objects ($F_s$) such as obstacles and barriers, local map information ($F_m$), and navigation information ($F_{nv}$).  Given the strong correlation between navigation and map, this work fuses navigation into the map information. The trajectory planning task can be expressed as: 
\vspace{-0.2cm}
\begin{equation}
    \mathcal{T}_{1:T_f} = f(F_e, F_a, F_s, F_m)
    \label{eq:prob}
\end{equation}
Where 
\begin{equation}
    \mathcal{T}_{1:T_f} = \{P_i\}_{i=1:T_f}, P_i=[x_i, y_i, \theta_i, v_i]
\end{equation}
represents the trajectory points for the next $T_f$ time steps, with each trajectory point $P_i$ consisting of the x and y coordinates, heading angle $\theta_i$, and velocity $v_i$ at the corresponding time step. The EMoE-Planner proposed in this paper serves as the function $f()$ in Equation~\ref{eq:prob}.
\paragraph*{Framework}
Figure~\ref{fig:framework} illustrates the framework of the proposed model. The model receives the ego vehicle state information $F_e\in R^{D_e}$, dynamic object information $F_a\in R^{N_a\times T_h \times D_a}$, static object information $F_s\in R^{N_s\times D_s}$, and map information $F_m\in R^{N_m\times L_p \times D_m}$. These inputs are processed through respective encoding modules to obtain D-dimensional features. The details of these input encoding modules will be elaborated in Section~\ref{sec:feature-encoders}. It is worth noting that the positional and heading angle information in the input data are relative values defined in the ego vehicle’s coordinate system. \par
After encoding, all processed features are concatenated and passed through the scene encoder module, which comprises $L_e$ layers of a normal Transformer encoder, resulting in the scene encoding representation $Q_S$. Subsequently, $Q_S$ is processed by the EMoE decoder module and the prediction decoder module to generate trajectory queries for the ego vehicle and surrounding vehicles, respectively. These trajectory queries are further processed by the prediction head and the trajectory head to output the predicted trajectories of surrounding vehicles and the planned trajectories for the ego vehicle. \par
The trajectories for the ego vehicle consist of multiple candidate trajectories, each accompanied by a probability. In the absence of a post-processing module, this work selects the trajectory with the highest probability as the final planned trajectory, which constitutes the output of the task described by Equation~\ref{eq:prob}. Although predicting trajectories is not the primary objective of this trajectory planning task, it serves as an auxiliary output to supervise the performance of the scene encoder.

\subsection{Feature Encoders}
\label{sec:feature-encoders}
\paragraph*{Ego Status Encoder(\textbf{EE}) \& Static Object Encoder(\textbf{ES})}
As Equation~\ref{eq:ee} shows, SDE~\cite{plantf} and multi-layer perceptrons (MLP) are employed to encode $F_e$ and $F_s$.  The ego status $F_e\in R^{D_e}$, where $D_e$ includes positional and heading information, is embedded into dimention $D$. Similarly, static object information $S\in R^{N_s\times D_s}$, where $N_s$ is the number of static objects and $D_s$ includes attributes such as position, heading, type, and shape, is also embedded into D for concatenation in later stages. $E_{ego}\in R^{D}$ and $E_{static}\in R^{N_s\times D}$ are encoded features of ego staus and static objects. \par
\vspace{-0.3cm}
\begin{equation}
    \label{eq:ee}
    E_{ego}=\text{SDE}(F_e),E_{static}=\text{MLP}(F_s)
\end{equation} 
\paragraph*{Agent Encoder(\textbf{EA}) \& Map Encoder(\textbf{EM})}
Unlike \textbf{EE} and \textbf{ES}, the encoding of agent and map information requires consideration of both temporal and spatial compression. Equation~\ref{eq:ae1} and~\ref{eq:ae2} shows the process of EA and EM. Raw input of agent is $F_a\in R^{N_a\times T_h\times D_a}$, where $N_a$ is the number of agents, $T_h$ the historical time steps, and $D_a$ encompasses attributes such as position, heading, velocity, shape, and type. The ecoded feature of dynamic agents is $E_{agent}\in R^{N_a\times D}$. Dimention $D_a$ is first embedded into dimention $D$ with Fourier embeddings (FoPE)~\cite{fope}, followed by a compression of $N_a$ to a singular dimension via MLP-Mixer~\cite{mixer-mlp}. Fope effectively capture periodic patterns in the data, facilitating effective temporal compression. Subsequently, the MLP-Mixer employs dual layers of MLPs to perform interactions both within and across the temporal dimension. Similarly, the map information $F_m\in R^{N_m\times L_p \times D_m}$ is also processed with two steps. But the first step is not with fope but a normal linear layer, which is better to handle spacial information. The ecoded feature of map infomation is $E_{map}\in R^{N_m \times D}$.
\begin{align} 
E_{agent}=\text{MLP-Mixer}(\text{FoPE}(F_a)) \label{eq:ae1}\\
E_{map}=\text{MLP-Mixer}(\text{Linear}(F_m)) \label{eq:ae2}
\end{align}

\subsection{Explicit MoE}
In autonomous driving trajectory planning, different scenarios require the ego vehicle to focus on specific infomation. In rule-based planners, tailoring specific rules for each type of scenarios simplifies them considerably compared to universal rules. Inspired by this, employing specialized experts for varying scenarios in a planning model can similarly simplify each expert's learning demands and enhance efficiency. To this end, we integrate a MoE structure into our model design.\par
\begin{figure*}[htbp]
  \centering 
  \includegraphics[width=0.85\textwidth]{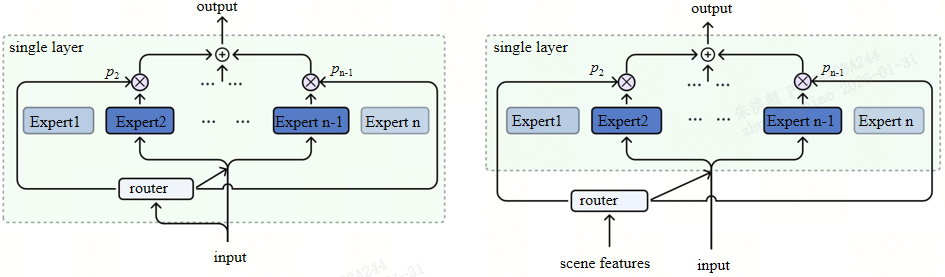} 
  \caption{Differences between vanilla MoE and EMoE} 
  \label{fig:moe}
  \vspace{-0.4cm}
\end{figure*}

Different from vanilla MoE, we propose Explicit MoE (EMoE), a new variant tailored for our task. Figure~\ref{fig:moe}(a) depicts a standard MoE model where at each layer of the decoder, a router selects top $N_E$ expert and their outputs are combined through weighted summation to produce the final output. However, this architecture has several limitations. Firstly, the router operates as a black box, whose outputs lack semantic meaning, making the selection logic neither interpretable nor monitorable. Secondly, it is challenging to ensure proper load balance among the different experts without extra design or loss. In LLMs, tasks are often ambiguous and lack clear delineation, making a semantic-agnostic router appropriate. This allows the model to learn implicit task classification and expert selection. However, for autonomous driving planning, scenarios can be explicitly divided into a finite set of sub-tasks. Thus, binding each expert to a specific sub-task emerges as a natural choice.\par
Figure~\ref{fig:moe}(b) illustrates EMoE. The architecture is distinguished by using only one router across all decoder layers. This router takes scene features as input, and its output, namely the identified scenario type, is shared by every layer. In this paper, the task scenarios can be explicitly divided into seven categories: turn left in junctions, go straight in junctions, turn right in junctions, go straight, roundabouts, U-turns, and others. The "others" category encompasses complex scenarios such as lane merging/diverging, lane number changes, and pull-over/starting, etc., which are difficult to classify reliably under conventional rules. All training data are pre-labeled using fixed rules. These data labels are then used to supervise the training of router. This enables the router to learn explicit and fixed routing rules, while each expert is dedicated to processing one specific scenario (e.g., Expert 3 only handles right-turns). Consequently, each expert specializes more effectively, improving individual performance—especially for data-scarce scenarios like U-turns and roundabouts—and boosting overall model capability. Notably, experts sharing the same index across all layers are assigned to the same scenario (e.g., Expert 3 in all layers process right-turn scenarios), unlike in vanilla MoE where experts in different layers are independent. This design also accelerates model convergence.\par
Furthermore, EMoE inherently enables load balancing via its data-centric design: each expert's workload is determined solely by the proportion of its assigned scenario in the dataset. This allows for targeted fine-tuning of experts by strategically adjusting the data distribution during training. \par

\subsection{Scene-specific query}
\label{sec:query}
We introduce scene-specific queries that incorporate prior knowledge at both decision and modality levels, enhancing the model’s focus on future areas relevant to intended actions. This approach not only stabilizes training but also simplifies the learning process. This concept aligns with human driving behaviors where attention is naturally directed towards areas pertinent to intended maneuvers, such as the left-front region when preparing to turn left. As discussed in section III-C, driving scenarios are categorized into seven types based on a set of predefined rules. Every training sample is then annotated with one of these scenario labels. By extracting all ground-truth trajectories from a given scenario type, we obtain a set that nearly covers all possible motion modalities of the ego vehicle over the 8-second horizon within that scenario. The endpoints of these trajectories represent the possible vehicle positions at 8 seconds. We then cluster all endpoints into $K_a$ classes using k-means, and the resulting cluster centroids serve as the anchor points for that scenario. Repeating this process for all seven scenario types yields a set of scene-specific anchor points, denoted as $G\in R^{7\times K_a\times 2}$ where the last two dimensions correspond to the x and y coordinates, respectively. Following scenario classification by the scene router, a specific set of target points $G_i\in R^{K_a\times 2}$ is selected for each scene, facilitating more precise model training and prediction accuracy. \par
\begin{figure}[htbp]
  \centering 
  \includegraphics[width=0.49\textwidth]{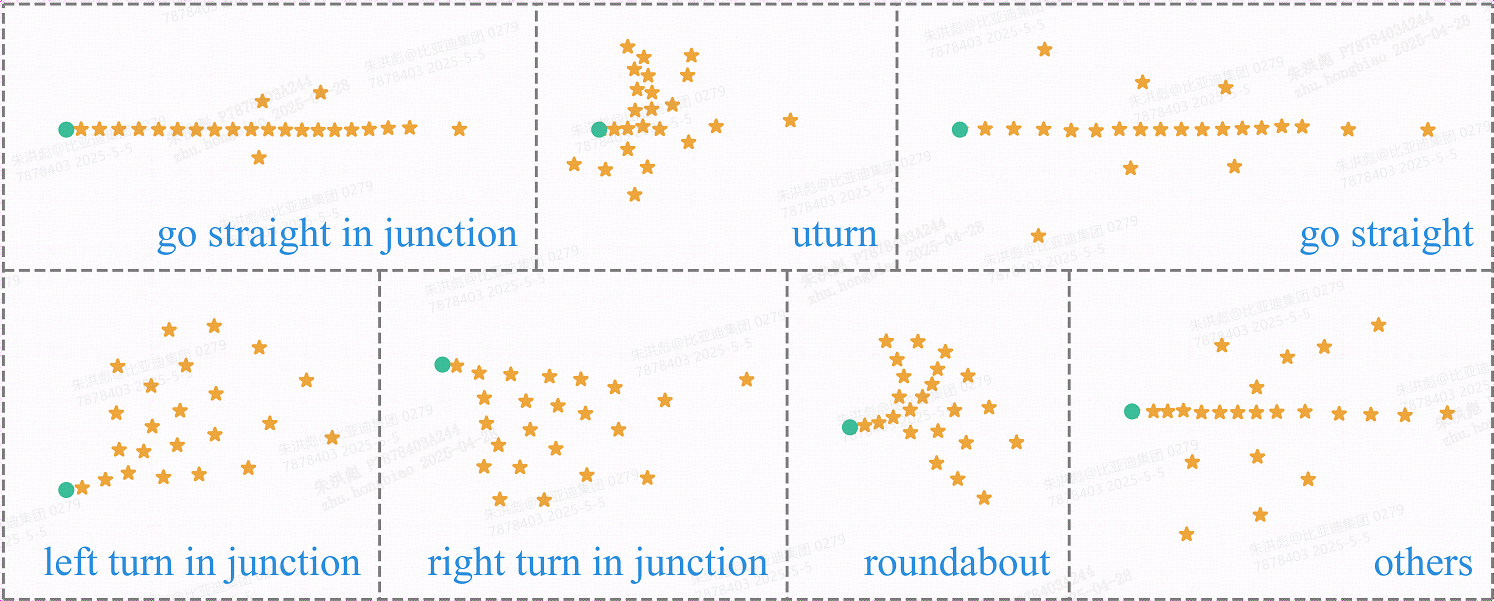} 
  \caption{Scene-specific anchors} 
  \label{fig:query} 
\end{figure}

The scene-specific query can be modeled with Equation~\ref{eq:query}.
\begin{equation}
    \label{eq:query}
    Q_{\text{mode}}=Q_{\text{learnable}}+\text{FoPE}(G_i)
\end{equation}
Where $Q_{\text{learnable}}\in R^{K_a\times D}$ is a learnable query. Each $Q_{\text{mode}}$ integrates prior knowledge across scenario and modality levels, targeting specific trajectory predictions within designated regions for each scenario. This architecture promotes trajectory generation with multiple modalities while ensuring concentrated attention, thereby stabilizing training and enhancing convergence speed. Moreover, the efficiency of our focused attention mechanism allows for a reduced number of queries (24 in our experiments), in contrast to traditional anchor-based models, leading to more effective region coverage and improved prediction quality. Figure~\ref{fig:query} shows the distribution of anchor points across the seven scenarios used in our study.\par

\subsection{Interaction-oriented loss and prediction}
\subsubsection{Interaction-oriented loss}
\paragraph*{Temporal-weighted regression loss} In existing imitation learning frameworks~\cite{pluto,plantf}, trajectory regression loss is typically calculated assuming equal importance across all trajectory points, with uniform weighting. However, real-world driving behaviors demonstrate a prioritization of immediate, short-term actions. Nearby obstacles, for instance, may necessitate evasive maneuvers and significantly influence the trajectory, whereas distant obstacles might not alter driving behavior significantly unless they come into closer proximity. This suggests the need for a differentiated approach to learning expert trajectories, where immediate responses to the near-future environment are prioritized over responses to distant conditions.\par
To better align with these observations, we introduce a temporal-weighted regression loss that scales the importance of each point in a trajectory based on its temporal closeness. The weight for the \textit{i}-th point is determined by the following formula.\vspace{-0.1cm}
\begin{equation}
    \label{eq:reg}
    W_{t_i}=e^{-k_R\times t_i}
\end{equation}
Where $k_R$ is a hyperparameter used to adjust the weight differences at different time steps, and $t_i$ the temporal distance into the future for the \textit{i}-th trajectory point.\par
A notable challenge in imitation learning is the issue of information leakage. This occurs when long-term behaviors in expert trajectories are responses to future events not yet present in the current observable scene. An overemphasis on learning these long-term behaviors can lead to trajectories that are counterintuitive. While the weighting mechanism described in Equation~\ref{eq:reg} does not completely eliminate the effects of information leakage, it substantially reduces its impact.

\paragraph*{Interaction-aware regression Loss}
In urban environments, the primary complexity arises from interactions between the ego vehicle and other agents, a topic not extensively explored in prior research. Our analysis of closed-loop simulation data from NuPlan reveals that the majority of failures are due to poor interactions between the ego vehicle and other vehicles. In response, this paper introduces an interaction-aware regression loss, which leverages critical interaction periods extracted from the ground truth data to enhance learning in interactive scenarios.\par
The method for extracting interaction intervals is as follows: traverse each time step $t$ and assess all trajectory points of other agents for potential collisions based on their shapes and orientation angles. Subsequently, using the results from the collision checks, we determine the beginning and ending of interaction intervals, $t_{in}, t_{out} \in [1:T_f]$. It is essential to clarify that the collisions mentioned herein do not pertain to actual physical interactions between vehicles but rather to instances where two vehicles occupy the same spatial location at different time steps. Furthermore, interaction labels (overtake/yield) are assigned based on the temporal sequence in which the vehicles arrive at the location. Specifically, if the ego vehicle at time $t_{e}$ “collides” with the \textit{i}-th vehicle at time $t_{a}$, and $t_{e}<t_{a}$ (i.e., the ego vehicle passes through first), then $t_{in}=t_{e}$, and the interacting vehicle is labeled as "overtake".\par 

For the \textit{i}-th agent, the process of $t_{\text{in}}^i$ and $ t_{\text{out}}^i$ is shown in Equation~\ref{eq:tin}-\ref{eq:tout}. The interaction interval is defined as the union of all ($t_{\text{in}}^i$,$t_{\text{out}}^i$) intervals, as shown in Equation~\ref{eq:all}.
\begin{equation}
\begin{split}
    t_{\text{in}}^i = \min \bigl\{ t_{a,i} &\mid \exists t_e \in \{1, \ldots, T_f\}, \\
    &\quad \| y_{a,i}^{t_{a,i}} - y_{e}^{t_e} \| < d_{\text{thr}} \bigr\}
\end{split}
\label{eq:tin}
\end{equation}
\begin{equation}
\begin{split}
    t_{\text{out}}^i = \max \bigl\{ t_{a,i} &\mid \exists t_e \in \{1, \ldots, T_f\}, \\
    &\quad \| y_{a,i}^{t_{a,i}} - y_{e}^{t_e} \| < d_{\text{thr}} \bigr\}
\end{split}
\label{eq:tout}
\end{equation}
\begin{equation}
    \label{eq:all}
    (t_{\text{in}}, t_{\text{out}}) = \bigcup_{i} (t_{\text{in}}^i, t_{\text{out}}^i)
    \vspace{-0.2cm}
\end{equation}
where $t_{a,i}$ and $y_{a,i}^{t_{a,i}}$ denote the time step and position of the \textit{i}-th agent, $t_e$ and $y_{e}^{t_e}$ denote the time step and position of the ego vehicle, $d_{\text{thr}}$ is a dynamic threshold at different time steps, depending on the agent size.\par
Based on the temporal-weighted regression loss in Equation~\ref{eq:reg},  we set the weights of the ego trajectory points within the interaction intervals to 1 to highlight their importance.
\begin{equation}
    \label{eq:kr}
    W_t=\begin{cases}
    1 & t_{\text{in}} \leq t \leq t_{\text{out}} \\
    e^{-k_Rt} & \text{else}
    \end{cases}
\end{equation}
The final regression loss can be calculated with Equation~\ref{eq:regloss}. 
\begin{equation}
    \label{eq:regloss}
    \mathcal{L}_{reg}=L1(T_{1:T_f}, T_{1:T_f}^{gt})\times W_{1:T_f}
\end{equation}
The overall loss of our model is shown in Equation~\ref{eq:loss}.
\begin{equation}
    \label{eq:loss}
    \mathcal{L} = w_1\mathcal{L}_{reg} + w_2\mathcal{L}_{cls} + w_3\mathcal{L}_{col} + w_4\mathcal{L}_{pred} +w_5\mathcal{L}_{router}
\end{equation}
where $w_1$,$w_2$,$w_3$, $w_4$ and  $w_5$ are weights of each loss. $\mathcal{L}_{cls}$, $\mathcal{L}_{col}$ and $\mathcal{L}_{pred}$ are imitation class loss, collision loss and prediction loss. These losses are similar to those in existing work~\cite{pluto,plantf}, thus we will not introduce them in detail. $\mathcal{L}_{router}$ is the loss of scene router mentioned in section 3.3. It can be calculated with the following equation.
\begin{equation}
    \label{eq:router-loss} 
    \mathcal{L}_{router} = \text{CrossEntropy}(\tau, \tau^{gt})
\end{equation}
where, $\tau$ is the output of scene router and $\tau^{gt}$ is the one-hot distribution derived from the ground truth label of the scenario. 

\subsubsection{Interaction-oriented prediciton}
The discourse in Section~\ref{sec:related-work} reveals that prevalent models typically overlook the dynamic interplay between the ego vehicle and other vehicles in trajectory prediction. Recognizing the real-world implications of such interactions, this study integrates the interaction of the ego vehicle’s planned trajectory with those of other vehicles into the prediction framework. As illustrated in Figure~\ref{fig:framework}, this is achieved by introducing a Pred Decoder module. This module synthesizes the agent encoding from the Scene Encoder with the ego vehicle trajectory features from the MoE Decoder via a cross-attention mechanism, facilitating interaction.\par
While these prediction outcomes do not directly feedback into the decoder or planning module, the interaction fosters enhanced prediction accuracy. This improvement indirectly refines the encoding performance of the Scene Encoder and Feature Encoders, thereby augmenting the overall precision of the trajectory planning process. 

\section{Experiments}
\subsection{Experimental Setup}
\paragraph*{Dataset}
The training and evaluation of the model were conducted using the NuPlan~\cite{nuplan} dataset. From this dataset, 3 million samples, spread across various scenarios, constituted the training sample pool. During preprocessing, samples were categorized into seven scenario types as outlined in Section~\ref{sec:query}. However, not all samples were utilized during the training process. Instead, the size of the training dataset was regulated by controlling the number of each scenario, detailed in Section~\ref{sec:results}.

\paragraph*{Benchmark and Metrics}
The model was evaluated using closed-loop non-reactive (NR) simulations with Val14~\cite{imitation1}, Test14, and Test14-hard~\cite{plantf} benchmarks. Although Nuplan offers reactive simulations, these are not considered robust due to their simplistic rule-based interactions. The industry uses non-reactive obstacle-based interactions and assigns the main responsibility for collisions, as done in Nuplan’s NR simulations. Therefore, we only use NR as the standard for evaluation. The evaluation metrics are the same with those in~\cite{nuplan-bench} and~\cite{nuplan2022}. For a fair and consistent comparison with existing works, all evaluation scores are scaled by 100 compared to official scores. Detailed information is in the supplementary materails.
\paragraph*{Baselines}
The EMoE-Planner was benchmarked against state-of-the-art (SOTA) algorithms categorized into three groups~\cite{pluto}: rule-based, learning-based and hybird. Algorithms like Pluto and DiffusionPlanner, which integrate rule-based post-processing following a learning-based approach, are classified as hybrid with post-processing while as learning-based without post-processing. Below is the list of baseline algorithms used for comparison. \par 
\begin{itemize} 
    \item IDM~\cite{idm}-- A rule-based trajectory-following method commonly used in planning simulations.
    \item PDM-Closed~\cite{imitation1} -- The method uses a common MLP, predicting trajectory with the centerline extracted through IDM and the history of the ego vehicle.
   \item GameFormer\cite{gameformer} -- The model enhances prediction accuracy for both the ego and surrounding vehicles by modeling their interactions based on game theory.
    \item UrbanDrive~\cite{urbandrive} -- A imitation-learning-based method which train a policy network in closed-loop employing policy gradients. 
    \item PlanTF~\cite{plantf} -- A method which applying transformer architecture to imitation learning. 
    \item Pluto~\cite{pluto} -- An imitaion learning method with more auxiliary losses and data augmentation.
    \item DiffusionPlanner~\cite{diffusionplanner} -- The method achieves the SOTA performance by employing the diffusion model and classifier guidance.
\end{itemize}

\paragraph*{Implementation Details}
We trained all models on eight Nvidia H800 graphics cards, each with 80GB of VRAM, setting the batch size to 32 and training for 35 epochs. The training time for one million data is approximately 16 hours.  The parameters used for training are listed in Table~\ref{tab:parameters}.
\begin{table}[!h]
  \caption{Parameters Used in EMoE-Planner}
  \label{tab:parameters}
  \centering
  \setlength{\tabcolsep}{2.9pt}
  \begin{tabular}{llllll}
    \hline
    Name     & Description     & Values & Name & Description & Values \\
    \hline
    lr & learning rate  & 1e-3 & $L_e$ & Num. scene encoder layers & 4\\
    $T_f$ & Future Steps & 80 & $k_R$ & Weight decay factor & 0.02\\
    $T_h$ & History Steps & 20 & $L_d$ & Num. decoder layers & 4\\
    $N_E$ & Num. experts & 7 & $D$   & Dimention of features & 128\\
    $K_a$ & Num. anchors & 24 & $D_h$ & Hidden dim of expert & 512\\
    \hline
  \end{tabular}
  \vspace{-0.3cm}
\end{table}

\subsection{Results}
\label{sec:results}
We assess the performance of all baseline algorithms and EMoE-Planner in NR simulations across Val14, Test14, and Test14-hard. Table~\ref{tab:results} presents the evaluation outcomes while Table~\ref{tab:metrics} details the metric results for the top two baseline methods and our method.
\begin{table}[!h]
  \caption{Closed-loop Results. Higher scores indicate better performance.}
  \label{tab:results}
  \centering
  \setlength{\tabcolsep}{3pt}
  \begin{tabular}{lllll}
    \hline
    Type & Planners & Val14 & Test14 & Test14-hard \\
    \hline
    Expert & Log-replay & 93.53 & 94.03 & 85.96 \\
    \hline
    \multirow{5}{*}{\makecell[l]{Rule-based \\ \& hybrid}}   
      & IDM & 79.31 & 70.39 & 56.15 \\
      & PDM-Closed & 93.08 & 90.05 & 65.08 \\
      & GameFormer & 79.94 & 83.88 & 68.70 \\
      & Pluto w/ post. & 92.88 & 92.23 & 80.08 \\
      & DiffusionPlanner w/ post. & 94.26 & \textbf{94.80} & 78.87 \\
      & EMoE-Planner w/ post. (Ours) & \textbf{94.35} & 91.45 & \textbf{80.96} \\
    \cline{1-5}  
    \multirow{6}{*}{\makecell[l]{Learning- \\ based}} 
      & UrbanDriver & 67.72 & 63.27 & 50.40 \\
      & PlanTF & 85.30 & 86.48 & 69.70 \\
      & Pluto w/o post. & 89.04 & 89.90 & 70.03 \\
      & DiffusionPlanner w/o post. & 89.87 & 89.19 & 75.99 \\
      & EMoE-Planner w/o post. (Ours) & \textbf{93.06} & \textbf{90.67} & \textbf{80.12} \\
    \hline
  \end{tabular}
\end{table}

\begin{table*}[htbp]
  \vspace{-0.2cm}
  \caption{Metrics Results.}
  \label{tab:metrics}
  \centering
  \begin{tabular}{llllllll}
    \hline
    Planners & Score & Collisions & TTC & Drivable & Comfort & Progress & Speed \\
    \hline
    Pluto w/o post. & 89.04 & 96.18 & 93.28 & 98.53 & 96.41 & 89.56 & \textbf{98.13} \\
    DiffusionPlanner w/o post. & 89.87 & 97.03 & 94.40 & 98.74 & 96.89 & 91.20 & 98.11 \\
    EMoE-Planner w/o post. (Ours) & \textbf{93.06} & \textbf{97.98} & \textbf{94.77} & \textbf{99.54} & \textbf{98.16} & \textbf{92.35} & 97.87 \\
    \hline
  \end{tabular}
  \vspace{-0.3cm}
\end{table*}
According to the results in Table~\ref{tab:results}, our model outperforms all other learning-based methods across all testing scenarios, particularly excelling in the val14 and test14-hard scenarios where it exceeds the state-of-the-art (SOTA) results by a minimum of 3 points. In the val14 scenarios, the performance of the EMoE-Planner closely approaches that of expert scores. Among rule-based methods, when implementing the same post-processing technique used in Pluto and DiffusionPlanner with our EMoE-Planner, it achieves the highest performance in the val14 and test14-hard scenarios, albeit slightly lower than DiffusionPlanner in the test14 scenario. It is noteworthy that the performance of EMoE-Planner with and without post-processing shows slight differences, suggesting that the trajectories produced by the model are of high quality, potentially diminishing the necessity for post-processing, aligning with our objectives. Additionally, the EMoE-Planner, without post-processing, scores only 0.02 points lower than PDM-Closed, which is a rule-based but state-of-the-art algorithm in the NuPlan Challenge, on the val14 test set. On all other datasets, it outperforms PDM-Closed. To our knowledge, we are the first pure learning model to achieve superior performance in closed-loop simulations compared to rule-based algorithms on the NuPlan dataset. In comparison, only when post-processing is applied do the Pluto and Diffusion Planners achieve scores higher than those of PDM-Closed. Furthermore, on the Test14-hard dataset, EMoE-Planner significantly outperforms PDM-Closed. This indicates that rule-based algorithms struggle to handle complex scenarios, while our model continues to perform well in these situations.\par

As indicated in Table~\ref{tab:metrics}, our model generally outperforms both Pluto and DiffusionPlanner in all metrics except speed. The EMoE-Planner achieves a higher progress score, indicating superior traffic throughput efficiency, while maintaining a slightly lower speed metric, reflecting a more assertive yet rational driving style. This is balanced by notable improvements in safety-critical metrics: fewer collisions and a higher TTC, demonstrating enhanced safety and better risk anticipation. This suggests that the more assertive speed profile primarily occurs in low-risk scenarios. Overall, the trajectories generated by the EMoE-Planner are both faster and safer, closely resembling human driving performance.\par
One key reason for this balanced performance is the design of the Interaction-Oriented Loss, which explicitly accounts for dynamic interactions in scenarios like overtaking to generate trajectories that are neither overly conservative nor risky. Another reason is the combined effect of the EMoE architecture and scene-specific queries, which simplifies expert learning per scenario and focuses attention on critical spatial areas implicitly enhancing the overall trajectory quality and leading to an overall improvement in metric scores. \par

In the EMoE architecture, the distribution of scenario quantities in the training data is crucial for model performance. The second and fifth column in Table~\ref{tab:scenarios} show the scenario counts in the training dataset. ST-J, LT-J, RT-J means go straight, turn left and turn right in junctions. ST means go straight in non-junction areas. UT is u-turn and RA is roundabout. During training, we capped each scenario type at 200,000 from a pool of 3,000,000 samples as discussed in section 4.1. The first four scenario types reached this cap, while the last three did not due to insufficient data in the 3M training database, resulting in a total of 1,083,804 samples used for training. We conduct separate simulation assessments for each scenario type, with results presented in Table~\ref{tab:scenarios}. \par
\vspace{-0.2cm}
\begin{table}[htbp]
  \caption{Results of different scenarios.}
  \label{tab:scenarios}
  \centering
  \begin{tabular}{cccccc}
    \hline
    Scenario Type & Number & Score & Scenario Type & Number & Score\\
    ST-J & 200000 & 98.36 & LT-J & 200000 & 93.28 \\
    RT-J & 200000 & 93.60 & UT & 17050 & 85.20 \\
    ST & 200000 & 91.28 & RA & 105756 & 89.33 \\
    Others & 160998 & 90.10 & / & / & / \\
    \hline
  \end{tabular}
  \vspace{-0.3cm}
\end{table}
Two key insights can be derived from Table~\ref{tab:scenarios}. Firstly, the quantity of scenarios in the training dataset positively correlates with the expert model’s performance and resulting scores. For instance, scenarios like u-turns and roundabouts, which have fewer instances, exhibit lower scores. Secondly, the complexity of the scene rules affects the model’s learning efficacy, simpler rules, such as those for going straight at an intersection, facilitate better performance. This observation aligns with our expectations. Using scenario-based simulations allows us to identify underperforming scenario types, providing a basis for targeted data enhancement or model optimization in future research.

\subsection{Ablation Studies}
This section presents ablation studies of each component and the weith decay factor $k_R$ with the val14 dataset.
\paragraph*{Ablation of each componets}: We conduc ablation studies of three components: EMoE, scene-specific query(SSQ), and interaction-oriented loss and prediction(I-Loss). Table~\ref{tab:ablations} displays the scores following the removal of each module. Ablation experiments reveal that interaction-oriented loss and prediction significantly enhance performance, underscoring the importance of interaction. EMoE and scene-specific query also play crucial roles in boosting model performance and these two are used in conjunction. The scene-specific query guides the model’s focus to key scene areas, and EMoE processes that area’s information with a scene-adapted expert. Additionally, Table~\ref{tab:ablations} shows that factoring in interactions yields better predictions for surrounding agents.\par
\begin{figure}[htbp]
  \centering   \includegraphics[width=0.35\textwidth]{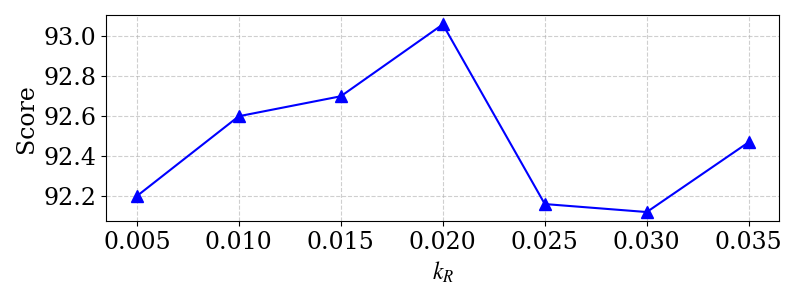} 
  \caption{Weight decay factor} 
  \label{fig:kr} 
  \vspace{-0.5cm}
\end{figure}

\vspace{-0.2cm}
\begin{table}[htbp]
  \caption{Ablation experiments.}
  \label{tab:ablations}
  \centering
  \begin{tabular}{llll}
    \hline
    Version & Score & AvgADE & AvgFDE\\
    \hline
    Base & 93.06 & 0.71 & 2.12\\
    w/o EMoE & 92.33 & 0.71 & 2.12\\
    w/o SSQ & 92.51 & 0.71 & 2.13\\
    w/o I-Loss & 91.44 & 0.73 & 2.20\\
    \hline
  \end{tabular}
  \vspace{-0.25cm}
\end{table}
\paragraph*{Weight decay factor $k_R$}: According to Equation~\ref{eq:kr} and ~\ref{eq:regloss}, $k_R$ has a significant impact on the overall regression loss. This set of experiments is conducted to test the influence of this parameter, and the results are presented in Fig~\ref{fig:kr}. According to Fig~\ref{fig:kr}, the optimal performance is achieved when $k_R$ is set to 0.02. A larger $k_R$ causes the model to over-focus on short-term trajectories, which degrades long-term planning performance. Conversely, decreasing $k_R$ results in insufficient attention to short-term interactions, leading to poorer performance in immediate scenarios.
\vspace{-0.2cm}

\subsection{Computation Efficiency}
This section presents a comparative analysis of the computational efficiency between EMoE-Planner, Pluto and DiffusionPlanner. The comparison is conducted across three key dimensions: the memory usage, the average inference latency, and the average floating point operations (FLOPs). The detailed results are summarized in Table~\ref{tab:efficiency}.
\begin{table}[htbp]
  \caption{Computation Efficiency.}
  \label{tab:efficiency}
  \centering
  \begin{tabular}{llll}
    \hline
    Planners & Memory(M) & FLOPs(G) & Latency(ms)\\
    \hline
    Pluto w/o post. & 4.24 & 0.82 & 20.75 \\
    DiffusionPlanner w/o post. & 6.04 & 1.54 & 77.44\\
    EMoE-Planner w/o post. & 8.91 & 2.42 & 36.27 \\
    \hline
  \end{tabular}
  \vspace{-0.3cm}
\end{table}
The EMoE-Planner incorporates 24 additional experts across its four decoder layers compared to other models, resulting in the largest memory usage. Furthermore, due to the optimized prediction decoder and the integration of scene-specific queries, the EMoE-Planner also exhibits the highest FLOPs. However, its inference latency remains competitive, which is only marginally higher than Pluto as only a single expert is active per inference. From an engineering perspective, the EMoE-Planner fully satisfies the real-time requirements for autonomous driving applications.

\subsection{Simulation Results}
\begin{figure*}[htbp]
  \vspace{-0.3cm}
  \centering   \includegraphics[width=0.92\textwidth]{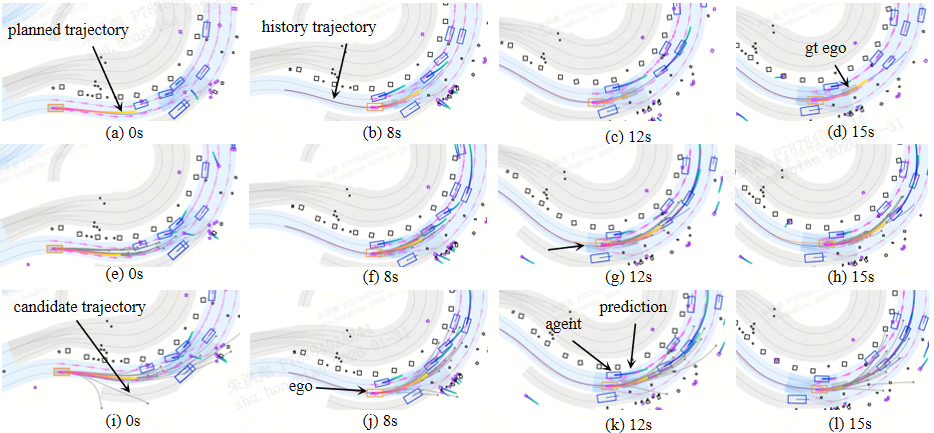} 
  \caption{Closed-loop simulation of roundabout. Subfigures (a) through (d) depict snapshots at 0s, 8s, 12s, and 15s for DiffusionPlanner, subfigures (e) through (h) represent snapshots at corresponding times for Pluto and subfigures(i) through (l) denote snapshots for EMoE-Planner. The ego vehicle is represented by an orange rectangle, while a gray rectangle denotes its ground truth position at the same timestamp. The planned trajectory is shown as a colored solid line, the historical trajectory in purple, and the ground-truth historical trajectory in blue. The candidate planned trajectories are displayed as gray lines.} 
  \label{fig:scenario1}
  \vspace{-0.3cm}
\end{figure*}

\begin{figure*}[htbp]
  \vspace{-0.3cm}
  \centering \includegraphics[width=0.90\textwidth]{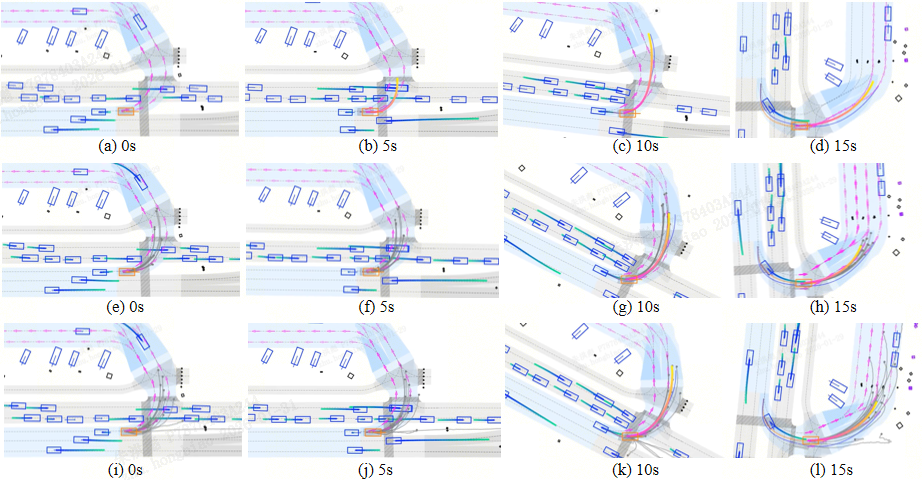} 
  \caption{Closed-loop simulation of u-turn scenario.} 
  \label{fig:scenario2}
  \vspace{-0.5cm}
\end{figure*}

Figures~\ref{fig:scenario1} and \ref{fig:scenario2} illustrate two of the least common scenario types among the seven defined categories: roundabout and u-turn maneuvers. We present snapshots at key time points for DiffusionPlanner, Pluto and EMoE-Planner. \par
In Figure~\ref{fig:scenario1}(b)(f)(j) at 8 seconds, Pluto collides with a left-side vehicle, while DiffusionPlanner maintains an uncomfortably close distance. In contrast, EMoE-Planner keeps a safe gap. By 12 seconds shown in Figure~\ref{fig:scenario1}(c)(g)(k), DiffusionPlanner collides with a vehicle on its right, and Pluto continues its collision on the left. EMoE-Planner, however, consistently maintains safe distances. Ultimately, only EMoE-Planner navigates the entire scenario without any collision. Comparing positions at 15s (Figure~\ref{fig:scenario1}(d)(h)(l)), EMoE-Planner travels the farthest, indicating the highest efficiency. This performance stems from two key design features of EMoE-Planner. First, its interaction-oriented loss explicitly accounts for interactions with other vehicles, which leads to significantly better collision avoidance. Second, the specialized expert dedicated to roundabout scenarios within the EMoE architecture allows the model to generate trajectories that are better adapted to the high-curvature geometry of roundabouts. This results in more reasonable and efficient paths compared to other planners.\par


In Figure~\ref{fig:scenario2}, all three planners safely navigate the U-turn. However, their positions at 15 seconds reveals that EMoE-Planner travels the farthest, even slightly exceeding the ground-truth trajectory by nearly one car length. Furthermore, according to historical trajectories at 15 seconds, DiffusionPlanner's path is excessively wide ("outer loop"), whereas EMoE-Planner selects a tighter, more efficient inner path. Similar to the roundabout case, the dedicated U-turn expert within the EMoE architecture enables better adaptation to the scenario's geometry, resulting in superior trajectories.\par
This section presents simulation results for two of the rarest and most challenging scenario types, where EMoE-Planner demonstrates significant performance improvements. In the supplementary material, we provide comparative results for the remaining five scenario types, including some particularly difficult cases and scenarios where EMoE-Planner still faces challenges. Across nearly all scenario types, EMoE-Planner outperforms both DiffusionPlanner and Pluto, especially in terms of safety and efficiency. \par


\section{Conculsion}
This paper introduces the EMoE-Planner, a novel framework tailored for enhancing trajectory planning in urban autonomous driving by utilizing scene information and interaction dynamics. The framework employs scene-specific queries generated through multiple scene anchors, focusing the model’s attention on relevant areas. Integration of a Mixture of Experts (MoE) with a shared scene router allows for precise matching of scene types with respective experts, improving the model’s responsiveness to varied scenarios and increasing the transparency of expert selection. Additionally, incorporating interactions between the ego vehicle and other agents into the loss function significantly improves trajectory scoring. Benchmark evaluations on several NuPlan datasets demonstrate that the EMoE-Planner surpasses state-of-the-art (SOTA) models by three points in closed-loop evaluations and closely aligns with the performance of post-processed SOTA methods, reducing reliance on rule-based post-processing. Furthermore, according to the simulation results, our model is the first pure learning model to achieve performance superior to rule-based algorithms (PDM-Closed) in NuPlan closed-loop simulations.\par
However, EMoE-Planner still has certain limitations. According to the results in Section IV(E) and supplementary material, our model currently underperforms in terms of traffic efficiency and in handling complex interactive scenarios compared to human drivers. Moreover, all evaluations in this work have been conducted within the nuPlan simulation environment. In future work, we aim to further refine the model to improve its performance in complex interactions and enhance its efficiency while maintaining safety. Most importantly, we plan to deploy and validate our model on real autonomous vehicles to advance its practical application. \par

\section*{Acknowledgments}
This work was supported by internal research funds from BYD Company Limited. No external funding was received. All authors are employees of BYD Company Limited. The research represents original work without conflicts of interest.

\bibliography{TITS}
\section{Biography Section}
\vspace{-1.2cm}
\begin{IEEEbiographynophoto}{Hongbiao Zhu}
received his Ph.D. from Harbin Institute of Technology in 2022 and currently serves as a Lead Algorithm Engineer at BYD Company Limited. His research focuses on trajectory planning, imitation learning and end-to-end models for autonomous driving.
\end{IEEEbiographynophoto}
\vspace{-30pt}
\begin{IEEEbiographynophoto}{Liulong Ma}
received his M.S. from Harbin Institute of Technology in 2019 and currently serves as a Lead Algorithm Engineer at BYD Company Limited. His
research focuses on interactive modeling and reinforcement learning in the fields of autonomous driving and robotics.
\end{IEEEbiographynophoto}
\vspace{-30pt}
\begin{IEEEbiographynophoto}{Xian Wu}
received his M.S. from Harbin Institute of Technology in 2023 and currently serves as a Senior Algorithm Engineer at BYD Company Limited. His research focuses on trajectory planning, imitation learning and end-to-end models for autonomous vehicle.
\end{IEEEbiographynophoto}
\vspace{-30pt}
\begin{IEEEbiographynophoto}{Xin Deng}
received his Ph.D. from Harbin Institute of Technology in 2023 and currently serves as a Senior Algorithm Engineer at BYD Company Limited. His research focuses on deep reinforcement learning and end-to-end models for autonomous vehicle.
\end{IEEEbiographynophoto}
\vspace{-30pt}
\begin{IEEEbiographynophoto}{Xiaoyao Liang}
received his M.S. from Southeast University in 2018 and currently serves as a Lead Algorithm Engineer at BYD Company Limited. His research focuses
on trajectory planning and decision making for autonomous driving.
\end{IEEEbiographynophoto}

\onecolumn 
\twocolumn %
\twocolumn[ 
\begin{@twocolumnfalse} 
\section*{\centering SUPPLEMENTARY MATERIAL} 
\end{@twocolumnfalse}
]
\subsection{Evaluation Metrics}
The evaluation metrics used in our study are the official benchmarks provided by the nuPlan platform~\cite{nuplan2022}, which have been widely adopted by existing research~\cite{pluto,diffusionplanner} utilizing the nuPlan dataset. The specific metrics are as follows:
\begin{itemize} 
    \item -- No ego at-fault collisions: A collision is defined as the event of ego’s bounding box intersecting another agent’s bounding box. Except for collisions that occur while the ego vehicle is stationary or when an active track hits the ego from the rear, all other collisions are considered at-fault. This metric score increases as the number and severity of at-fault collisions decrease. This metric is used to evaluate the safety of the trajectory.  
    \item -- Drivable area compliance: This metric identifies the frames when ego drives outside the drivable area. This metric is a binary criterion. If any frame within a clip is judged to be outside the drivable area, the score is 0. A score of 1 is only achieved if the trajectory remains within the drivable area for all frames. 
    \item Time to Collision (TTC) within bound: TTC is defined as the time required for ego and another track to collide if they continue at their present speed and heading. This metric is also boolean, 1 if TTC is higher than a minimum lower bound, for example 0.95s, else 0. TTC is indicative of driving caution, a higher TTC value signifies a more conservative driving style.
    \item -- Progress: This metric evaluates how much of the expert's path the ego vehicle has covered. It is calculated by comparing the distance traveled by the ego vehicle to the total length of the expert's route. A higher value for this metric directly indicates higher throughput efficiency of the planned trajectory. 
    \item -- Speed limit compliance: This metric evaluates if ego’s speed exceeds the associated speed limit in the map. The score for this metric is determined by considering both the duration and the severity of the overspeeding violation.
    \item -- Comfort: The comfort level is assessed by measuring the extreme values of critical kinematic parameters, encompassing longitudinal and lateral accelerations, yaw rate, yaw acceleration, and jerk.
\end{itemize}
Given the complexity involved in the detailed calculation of these metrics, the official nuPlan website~\cite{nuplan2022} remains the definitive source for complete information.

\subsection{Additional Simulation Results}
\begin{figure*}[htbp]
  \vspace{-0.2cm}
  \centering 
  \includegraphics[width=0.94\textwidth]{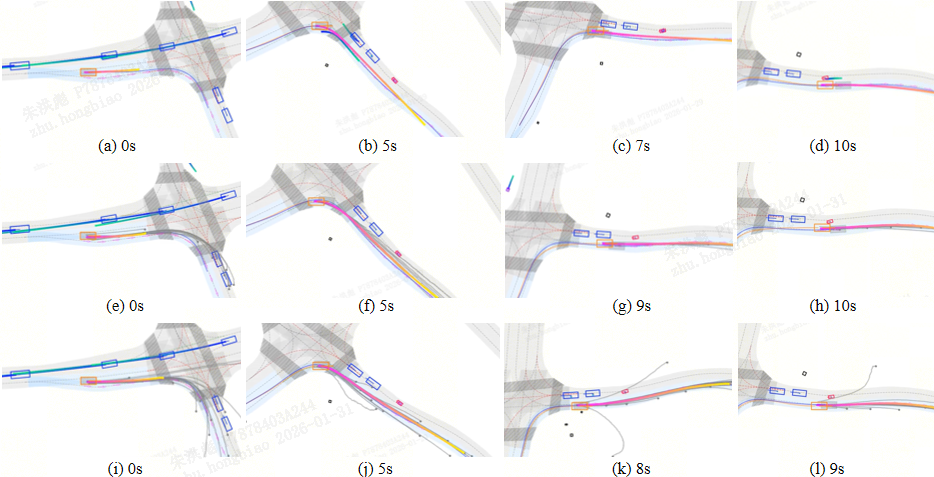} 
  \caption{Closed-loop simulation of right-turn scenario. Subfigures (a) through (d) depict snapshots at 0s, 5s, 7s, and 10s for DiffusionPlanner, subfigures (e) through (h) represent snapshots at 0s, 5s, 9s, and 10s for Pluto and subfigures(i) through (l) denote snapshots at 0s, 5s, 8s, and 9s for EMoE-Planner.} 
  \label{fig:scenario3}
\end{figure*}

\begin{figure*}[htbp]
  \vspace{-0.2cm}
  \centering 
  \includegraphics[width=0.94\textwidth]{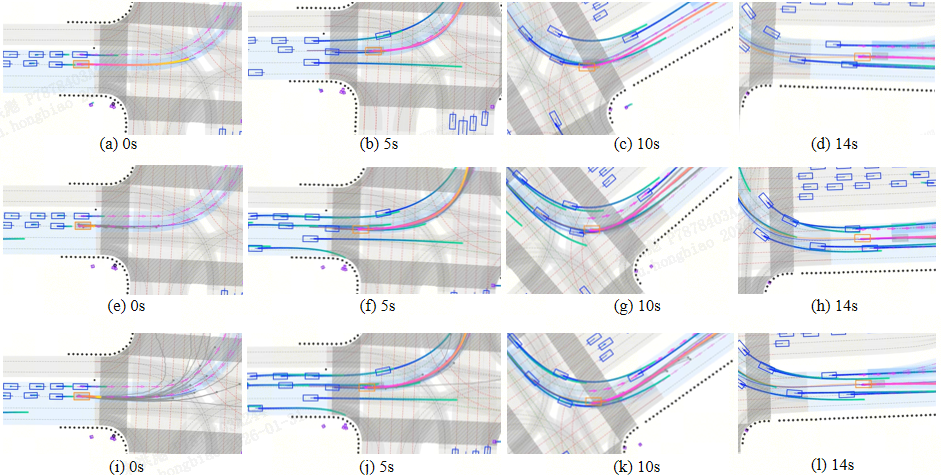} 
  \caption{Closed-loop simulation of left-turn scenario.} 
  \label{fig:scenario4}
  \vspace{-0.5cm}
\end{figure*}

\begin{figure*}[htbp]
  \vspace{-0.2cm}
  \centering 
  \includegraphics[width=0.99\textwidth]{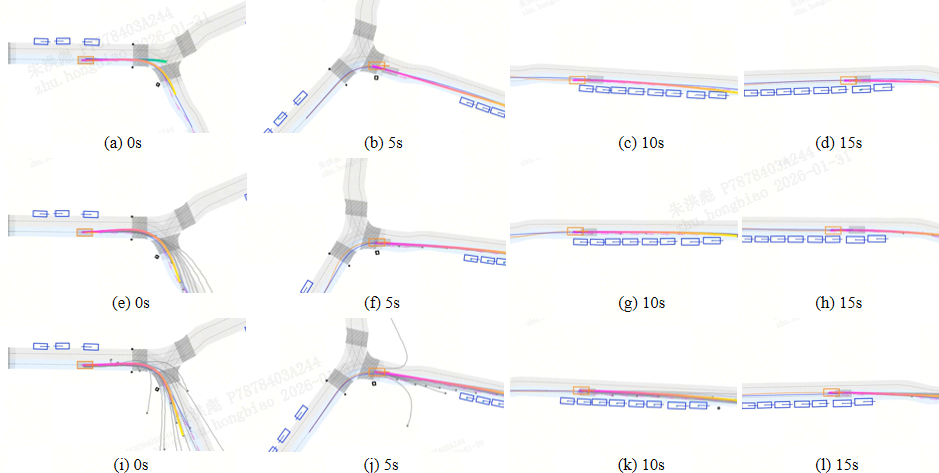} 
  \caption{Closed-loop simulation of going straight scenario.} 
  \label{fig:scenario5}
\end{figure*}

\begin{figure*}[htbp]
  \vspace{-0.2cm}
  \centering 
  \includegraphics[width=0.99\textwidth]{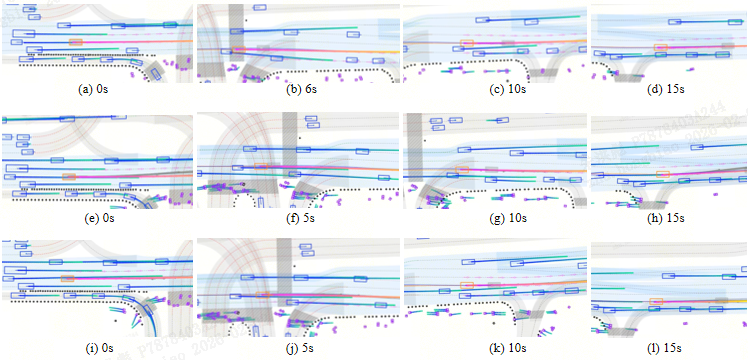} 
  \caption{Closed-loop simulation of going straight within a junction.} 
  \label{fig:scenario6}
  \vspace{-0.5cm}
\end{figure*}

\begin{figure*}[htbp]
  \vspace{-0.2cm}
  \centering 
  \includegraphics[width=0.99\textwidth]{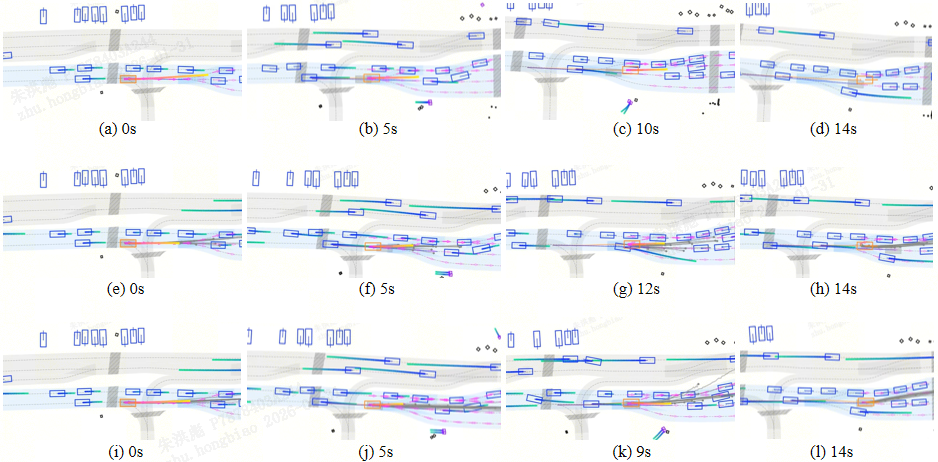} 
  \caption{Closed-loop simulation of other scenarios.} 
  \label{fig:scenario7}
  \vspace{-0.5cm}
\end{figure*}

Figure~\ref{fig:scenario3}-~\ref{fig:scenario7} present closed-loop simulation results for the five additional scenario types beyond those discussed in the main text. For each scenario, we captured screenshots at the critical timestep specific to each model, as their key decision points may differ. The primary challenge varies per scenario: left turns require balancing turning radius to avoid collisions with inner or outer lane vehicles; right turns must manage conflicts with oncoming cross-traffic; non-intersection going straight focuses on avoiding stationary vehicles during lane changes; going straight within intersections necessitates careful acceleration control to prevent rear-end or front-end collisions; and other complex scenarios like merging prioritize correct lane selection and dynamic collision risk management. \par
In Figure~\ref{fig:scenario3}, the DiffusionPlanner’s trajectory exhibits an excessively wide outer swing, resulting in a collision with a vehicle in the oncoming lane at 7s. Meanwhile, the PlutoPlanner passes dangerously close to an oncoming vehicle at the 9-second mark and subsequently collides with a red car in the opposing lane at 10 seconds. In contrast, the EMoE-Planner safely navigates through the intersection with a trajectory that closely aligns with the ground-truth path. Both in terms of efficiency and safety, EMoE-Planner performs the best in the right-turn scenario. \par
Figure~\ref{fig:scenario4} shows the result in left-turn scenario. The DiffusionPlanner’s trajectory exhibits an excessively wide outer loop as in Figure~\ref{fig:scenario3}, causing it to collide with a vehicle on the right during its return at the 14-second mark. Meanwhile, the PlutoPlanner suffers from inefficient steering, resulting in a rear-end collision by a following vehicle at 10 seconds. In contrast, the EMoE-Planner generates a trajectory nearly identical to the ground truth, completing the turn both safely and efficiently. \par

Figures~\ref{fig:scenario5} and~\ref{fig:scenario6} show the results for straight driving on a non-intersection road and going straight through an intersection, respectively. In the non-intersection scenario, both DiffusionPlanner and Pluto experience collisions at different time steps, while EMoE-Planner completes the maneuver safely. Moreover, based on the ego vehicle’s position at 15s, EMoE-Planner again achieves the highest efficiency. In Figure~\ref{fig:scenario6} the intersection straight-driving scenario, DiffusionPlanner and Pluto both exhibit slow starts, leading to inefficient progression and rear-end collisions by following vehicles at 6s and 10s, respectively. In contrast, EMoE‑Planner traverses the intersection without any collision. \par
As noted in the previous section, the no ego at-fault collisions metric does not penalize rear-end collisions. However, because EMoE-Planner’s interaction-oriented loss explicitly accounts for yielding and overtaking behaviors, rear-end interactions are also considered during optimization, enabling the model to effectively avoid such collisions. \par
Figure~\ref{fig:scenario7} presents the simulation results for a lane-splitting scenario, which falls under the "others" category and involves challenges in lane choice and lane-change timing. DiffusionPlanner changes lanes too early, resulting in a collision at the 10-second mark. Pluto, on the other hand, drives too slowly and is rear-ended by a following vehicle at 12 seconds. While EMoE-Planner’s trajectory closely aligns with the ground truth, it comes dangerously close to a vehicle on the left at 14 seconds, indicating a potential collision risk. Nevertheless, EMoE-Planner achieves the best overall performance among the three planners in this scenario. \par
\begin{figure*}[htbp]
  \vspace{-0.2cm}
  \centering 
  \includegraphics[width=0.99\textwidth]{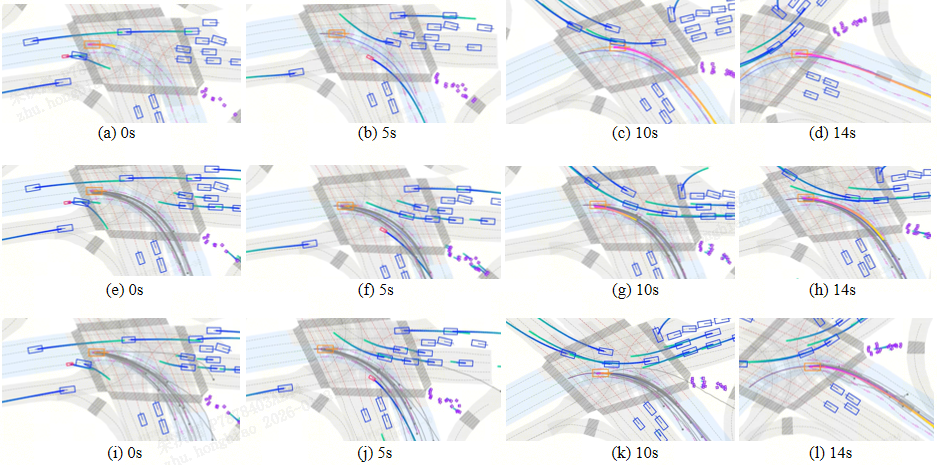} 
  \caption{Closed-loop simulation of a difficult case.} 
  \label{fig:scenario8}
  \vspace{-0.5cm}
\end{figure*}

The scenarios presented above are relatively common cases within each category. Figure~\ref{fig:scenario8} illustrates a more challenging right-turn scenario where the ego vehicle must navigate while multiple consecutive oncoming vehicles are executing left turns, creating a complex interaction zone with high collision risk. In this difficult setting, DiffusionPlanner again fails due to an excessively wide turning radius, resulting in a collision at 10 seconds. Conversely, PlutoPlanner adopts an overly conservative strategy, remaining almost entirely outside the interaction zone for the first 15 seconds and waiting for all oncoming traffic to clear before proceeding. EMoE-Planner successfully completes the right turn without collision. While its efficiency in this complex scenario is lower than that of the ground-truth trajectory, its overall behavior demonstrates a balanced approach: it remains cautious in the face of dense interactive traffic but continues to make progress without coming to a complete stop or moving excessively slowly, thereby maintaining a reasonable level of traffic throughput.
\begin{figure*}[htbp]
  \vspace{-0.2cm}
  \centering 
  \includegraphics[width=0.99\textwidth]{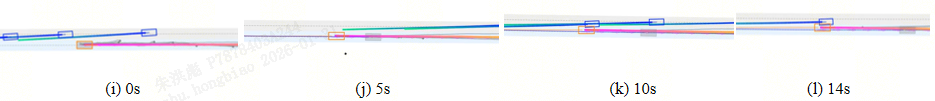} 
  \caption{Closed-loop simulation of a failure case.} 
  \label{fig:scenario9}
  \vspace{-0.5cm}
\end{figure*}
Finally, Figure~\ref{fig:scenario9} presents a case that the current EMoE-Planner still struggles to handle: a head-on encounter on a bidirectional single-lane road where an oncoming vehicle partially intrudes into the ego’s lane. Insufficient proactive avoidance capability leads to a collision in this scenario. Although such interactions are theoretically covered by the interaction-oriented loss, they are extremely rare in the training data, so the model has not effectively learned a robust avoidance policy. Furthermore, in this particular setting, the available evasion space is very limited, and any evasive maneuver risks driving outside the drivable area, thus, the model refrains from taking action. The optimal strategy, as demonstrated by the ground-truth trajectory, would be to accelerate and pass through the conflict zone before the oncoming vehicle fully intrudes. This case highlights that the ability of EMoE-Planner to balance interactive safety and traffic efficiency in highly constrained, adversarial scenarios still requires further improvement, which will be a central focus of our future work.
\end{document}